\documentclass[runningheads]{llncs}
\usepackage{eccv}
\usepackage{eccvabbrv}
\usepackage{graphicx}
\usepackage{booktabs}
\usepackage[accsupp]{axessibility}
\usepackage[pagebackref,breaklinks,colorlinks,citecolor=eccvblue]{hyperref}
\usepackage{orcidlink}
\usepackage{bm}
\usepackage{multirow}
\usepackage{colortbl}
\usepackage{bbding}
\usepackage{array}
\usepackage{tabularx}
\usepackage{algorithm}
\usepackage{algpseudocode}
\usepackage[misc]{ifsym}

\newcommand{\bI}{\mathbf{I}}

\newcommand{\bzero}{\mathbf{0}}

\newcommand{\bp}{\mathcal{P}}

\newcommand{\bz}{\mathbf{z}}

\newcommand{\bepsilon}{{\boldsymbol{\epsilon}}}
\newcommand{\bmu}{{\boldsymbol{\mu}}}

\definecolor{lightgray}{gray}{.9}

\newcolumntype{C}[1]{>{\PreserveBackslash\centering}p{#1}}

\makeatletter
\renewcommand\@fnsymbol[1]{\ensuremath{*}}
\makeatother

\begin{document}
\title{R3D-AD: Reconstruction via Diffusion for 3D Anomaly Detection}

\author{Zheyuan Zhou\inst{1 *} \and
Le Wang\inst{1}\thanks{Equal contribution.} \and
Naiyu Fang\inst{1, 2} \and \\
Zili Wang\inst{1} \and
Lemiao Qiu\inst{1} \and
Shuyou Zhang\inst{1}}

\authorrunning{Z.~Zhou et al.}

\institute{Zhejiang University \and
Nanyang Technological University\\
\ \\
\url{https://zhouzheyuan.github.io/r3d-ad}
}

\maketitle

\begin{abstract}
3D anomaly detection plays a crucial role in monitoring parts for localized inherent defects in precision manufacturing.
Embedding-based and reconstruction-based approaches are among the most popular and successful methods.
However, there are two major challenges to the practical application of the current approaches: 
1) the embedded models suffer the prohibitive computational and storage due to the memory bank structure; 
2) the reconstructive models based on the MAE mechanism fail to detect anomalies in the unmasked regions.
In this paper, we propose R3D-AD, reconstructing anomalous point clouds by diffusion model for precise 3D anomaly detection. 
Our approach capitalizes on the data distribution conversion of the diffusion process to entirely obscure the input's anomalous geometry.
It step-wisely learns a strict point-level displacement behavior, which methodically corrects the aberrant points. 
To increase the generalization of the model, we further present a novel 3D anomaly simulation strategy named Patch-Gen to generate realistic and diverse defect shapes, which narrows the domain gap between training and testing.
Our R3D-AD ensures a uniform spatial transformation, which allows straightforwardly generating anomaly results by distance comparison.
Extensive experiments show that our R3D-AD outperforms previous state-of-the-art methods, achieving 73.4\% Image-level AUROC on the Real3D-AD dataset and 74.9\% Image-level AUROC on the Anomaly-ShapeNet dataset with an exceptional efficiency.

\end{abstract}
\section{Introduction}
Anomaly detection aims to identify instances containing anomalies and to precisely locate the specific positions of defects.
This task is extensively applied across multiple fields and plays a crucial role in quality control within industrial production~\cite{patchcore}.
3D anomaly detection~\cite{real3dad} has emerged due to its intrinsic modality superior for avoiding blind spots in advanced processing and precision manufacturing.
However, the discrete and disordered data form of point clouds makes it more difficult to acquire features compared to images.
With the scarcity of anomalies, 3D anomaly detection also faces the problem of domain shift while only normal data are presented during training.
The presence of these issues underscores the necessity and urgency of devising an efficient framework for the 3D anomaly detection task.

\begin{figure}[t]
\centering
\begin{minipage}[t]{0.8\textwidth}
\includegraphics[width=\linewidth]{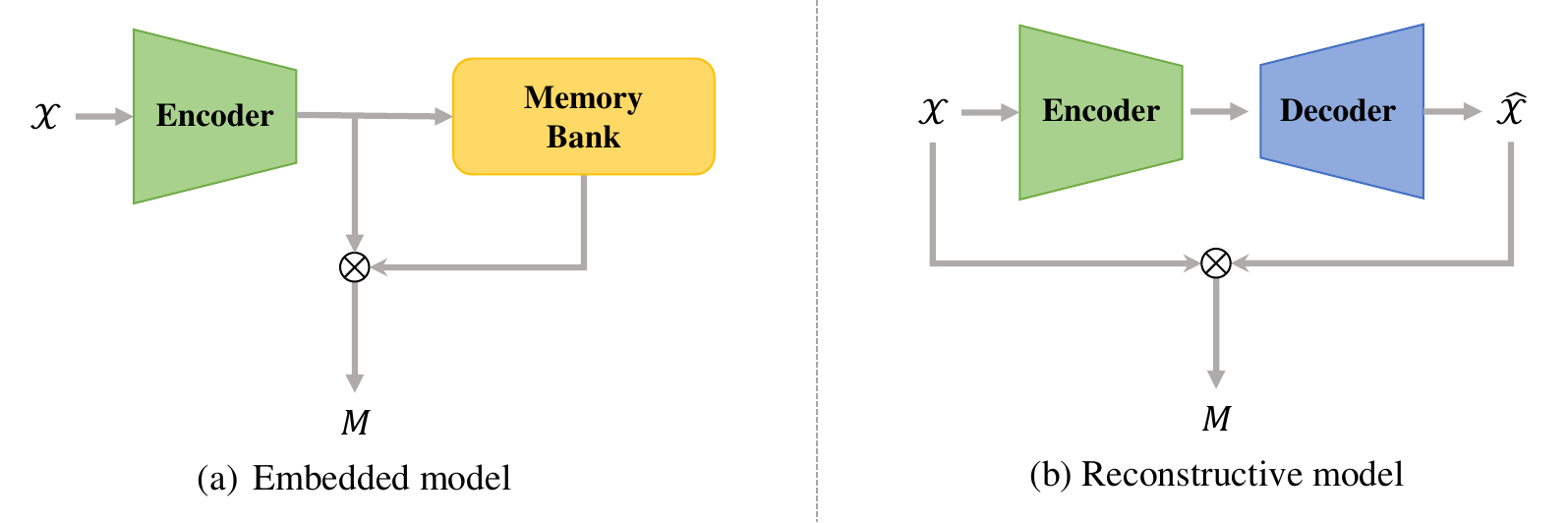}  
\end{minipage}
\caption{Comparison of architectures.
\textbf{(a)} Embedded model encodes the input $\mathcal{X}$ into features and stores them in the memory bank during training.
The anomaly map $\mathcal{M}$ is obtained by comparing the test features with all the features in the memory bank.
\textbf{(b)} Reconstructive model is trained by minimizing the loss between its input $\mathcal{X}$ and the output $\widehat{\mathcal{X}}$.
The anomaly map $\mathcal{M}$ is obtained by comparing the test phase input with its corresponding reconstruction target. 
}
\label{fig:archi}
\end{figure}

Similar to traditional 2D anomaly detection~\cite{patchcore, zavrtanik2021reconstruction}, current 3D anomaly detection can be primarily categorized into embedding-based and reconstruction-based, as illustrated in Fig.~\ref{fig:archi}.
The embedding-based methods involve mapping features extracted with a pre-trained encoder onto a normal distribution for learning.
Distributions that do not fall within the interval are classified as anomalies.
Most existing 3D anomaly detection methods are based on a memory bank mechanism~\cite{btf, m3dm, real3dad, cpmf}, which stores some representative features during the training phase to implicitly construct a feature distribution. 
In the testing phase, the presence of anomalies is determined by calculating the Euclidean distance between the input test object and all template point clouds stored in memory.
The reconstruction-based methods train a network capable of accurately reconstructing normal point clouds, under the presumption that anomalous point clouds will not be effectively reconstructed since they are not included during training.
The anomaly map is produced through the comparison of discrepancies between the input point cloud and its reconstruction.
IMRNet~\cite{shapenetad} employs PointMAE~\cite{pointmae} to reconstruct the input in several iterations, getting the final anomaly map by calculating the explicit spatial coordinate differences and implicit deep feature differences of the point cloud, respectively.

However, existing methods face two key issues, high resource cost and irreparable reconstruction.
Firstly, methods based on the memory bank~\cite{btf, m3dm, real3dad, cpmf} store all features from the training phase, each test point cloud needs to be compared with all samples in the memory bank, significantly increasing memory overhead and inference time costs. 
This makes such methods almost inapplicable in real industrial production lines due to their inefficiency.
Secondly, masked autoencoder (MAE) mechanism~\cite{mae, simmim, pointmae} only reconstructs the masked portions of the input, defects within unmasked portions may be preserved.
This contradicts the fundamental assumption of detecting anomalies by comparing the original defect-containing point cloud with a reconstructed anomaly-free version.
These methods inevitably lead to incorrect reconstructions, undermining their effectiveness in accurately localizing defects.

We propose R3D-AD, a novel 3D anomaly detection method that does not suffer from the space burden and time endurance in memory-based embedded models nor the anomaly unmasking probability in the MAE-based reconstructive models. 
In contrast to PointMAE, one of our key insights is to perform undifferentiated masking for 3D objects via the noise diffusion mechanism, which maximizes the preservation of anomaly-free shapes and reconstructs abnormal regions. 
In the reparameterized diffusion process, one-step full mask and reconstruction are achieved by converting the point cloud distribution, instead of the multiple iterative method~\cite{shapenetad}. 
We hypothesize that anomaly detection verifies the gap between the reconstructed shapes and the positive samples by learning point movement. 
Specifically, for input models with arbitrary anomalies, we encode them as latent shape embeddings as decoding conditions and explicitly control the point cloud reconstruction process by step-wise displacements (SWD) decoding. 
The shape embedding harbors abundant global features and makes it easier to train the network without dwelling on the introduction of local anomaly details. 
Another key to our approach is to implement a controllable method of point-wise displacement during the diffusion process to refine the point cloud deformation iteratively. 
We propose to inject latent shape embedding into each step of the inverse denoising process, which drives the anomalous regions to converge to a smooth surface.
We further adopted a 3D anomaly simulation strategy Patch-Gen to address the limitations of the dataset, which generates abundant defectives by producing spatial irregularity that is faithful to the real scene, including bulges, sinks, etc. 
This point cloud data augmentation encourages the self-supervised model to reconstruct more realistic anomaly-free shapes when facing the actual anomaly.

To the best of our knowledge, this is the very first attempt at exploring diffusion in reconstruction-based 3D anomaly detection.
Our main contributions are summarized as follows:
\textbf{(i)}
We introduce a novel framework, termed R3D-AD, which performs a one-step full mask and anomaly-free reconstruction for fast and accurate 3D anomaly detection.
\textbf{(ii)}
We propose to learn the step-wise displacement in the reverse diffusion process to explicitly control the reconstruction of anomalous shapes.
\textbf{(iii)}
We introduce a 3D anomaly simulation strategy named Patch-Gen to address the limitation of the data anomaly patterns and improve the reconstruction performance in a supervised setting.
\textbf{(iv)}
Extensive experiments demonstrate that our R3D-AD has achieved state-of-the-art performance on both Real3D-AD and Anomaly-ShapeNet datasets.
\section{Related work}

\subsection{2D Anomaly Detection}
Anomaly detection has received increasing attention from researchers in recent years, and many new methods have been proposed to address the problem.
Flow-based methods~\cite{differnet, cflow, uflow, fastflow} use learned distributions and flow's bijective properties to spot defects, while 
Memory-based approaches~\cite{patchcore, fapm, pni} gauge anomaly scores by contrasting test sample features with memory bank-stored norms.
Reconstruction-based models~\cite{ae_ssim, zavrtanik2021reconstruction, draem} flag anomalies by comparing inputs to their online reconstructions.
Recent works~\cite{cutpaste, nsa, anomalydiffusion, realnet} augment the anomaly detection datasets with generated synthetic anomalies to compensate for the negative example scarcity problem.

\subsection{3D Anomaly Detection}
This field lags behind the development of 2D anomaly detection since 3D data are harder to obtain, while point cloud data are sparser and contain more noise than image data.
BTF~\cite{btf} integration of handcrafted 3D descriptors with classic 2D method PatchCore~\cite{patchcore}, constructing a basic framework for 3D anomaly detection. 
M3DM~\cite{m3dm} advances the field by separately analyzing features from point clouds and RGB images, then merging these for improved decision-making.
CPMF~\cite{cpmf} converts point clouds into two-dimensional images from multiple angles, extracting additional features from these images with a pre-trained network, and enhancing detection capabilities through information fusion.
Reg3D-AD~\cite{real3dad} develops a registration-based method, the RANSAC algorithm was used to align each sample before comparing it to the stored template during the test phase.
IMRNet~\cite{shapenetad} trains a PointMAE~\cite{pointmae} to reconstruct anomaly-free samples and identifies anomalies by juxtaposing the reconstructed point cloud against the initial input.
Many of these use memory banks to store the features of the training samples or require multiple iterations to restore points. 
Unlike previous methods, our approach requires only one step of reconstruction and has significant advantages in both time and space efficiency.

\subsection{Diffusion Models} 
Diffusion models have proven their effectiveness in several generative tasks, such as image generation~\cite{imagen}, speech generation~\cite{diffwave}, and video generation~\cite{imagenv}.
Denoising Diffusion Probabilistic Models (DDPMs)~\cite{ddpm, sde, ddim} employ a forward noising mechanism, incrementally integrating Gaussian noise into images, alongside a reverse process meticulously trained to counteract the forward mechanism.
Denoise AD~\cite{denoisead} conducts DDPM for reconstructing within the features space, generating images that contain less noise.
In recent years, many studies~\cite{dit3d, meshdiffusion, diffcomplete, diffusion-sdf} have attempted to use the diffusion model to explore the 3D reconstruction task. 
DPM~\cite{dpm} incorporates a shape latent variable to encapsulate the geometric intricacies of 3D shapes, it distinctively models this variable's distribution utilizing Normalizing Flows~\cite{nf, rnvp}.
PVD~\cite{pvd} utilizes PVCNNs~\cite{pvcnn} for the point-voxel representation of 3D shapes and integrates structured locality into point clouds. 
This innovative approach leverages the strengths of both point and voxel representations, optimizing the model's ability to capture the intricate spatial hierarchies and local geometries within 3D objects.
Since diffusion-based reconstruction recovers the target shape from complete noise, the dilemma of reconstructing only the masked region in the MAE~\cite{mae} mechanism does not exist. 
\section{Method}
\subsection{Overview}
\begin{figure}[t]
\centering
\includegraphics[width=\linewidth]{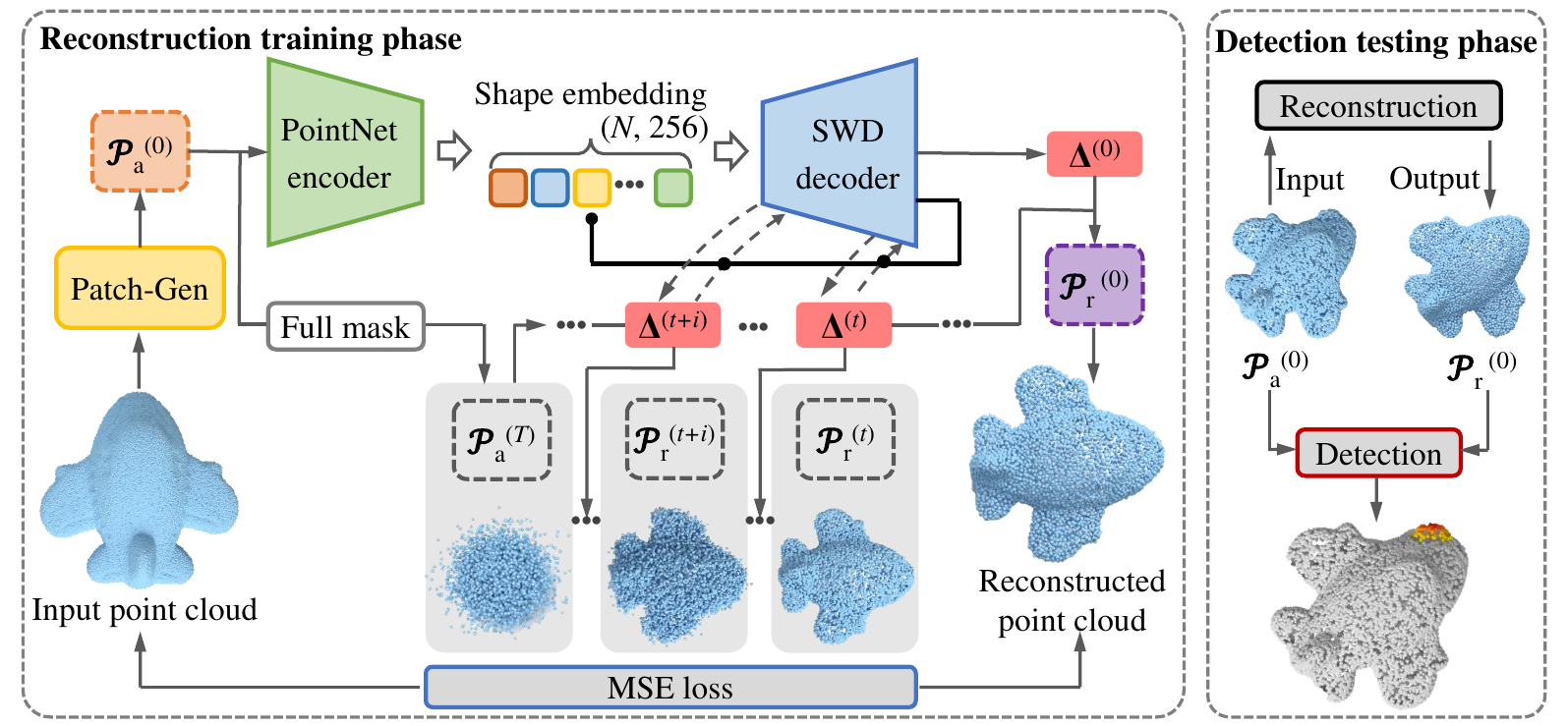}
\caption{Overall architecture of R3D-AD for shape reconstruction and anomaly detection of point cloud objects. 
\textbf{Reconstruction training phase}: The simulated anomalous  $\mathcal{P}_\text{a}^{(0)}$ is generated resort to Patch-Gen from the input point cloud.
It further fully masked as $\mathcal{P}_\text{a}^{(T)}$ while also encoded to latent shape embedding. 
The SWD decoder then explicitly reconstructs the anomaly-free object $\mathcal{P}_\text{r}^{(0)}$ with consistent spatial transform by conditionally generating point-level displacements $\mathrm{\Delta} ^{(t)}$ at each step of the inverse process. 
\textbf{Detection testing phase}: The test point cloud $\mathcal{P}_\text{a}^{(0)}$ is reconstructed to $\mathcal{P}_\text{r}^{(0)}$ with normal shape, and compared at a distance level to detect the anomalous region.}
\label{fig:framework}
\end{figure}

We model the anomaly detection problem as mapping an anomalous point cloud ${\mathcal{P}}_\text{a} \in \mathbb{R}^{N \times 3} $ to a positive shape with which it is aligned. 
The framework of R3D-AD is shown in Fig.~\ref{fig:framework}, where the simulated anomalous shapes are reconstructed in a self-supervised setting in the training phase and then compared with the original input to detect anomalies. 
The reconstructed anomaly-free model is aligned with the input, thus allowing direct computation of anomaly scores and segmentation of anomalous regions by conditioned distance functions. 
Simultaneously, the anomaly simulation strategy faithfully generates realistic defects and randomly synthesizes diverse anomaly shapes on normal samples, improving the generalization ability of the network in the case of limited anomaly samples.

\subsection{Preliminary of denoising diffusion probabilistic models}
A DDPM is inspired by the thermal diffusion process in an evolving thermodynamic system, which consists of a diffusion process and a reverse process.

The forward Markovian process gradually adds Gaussian noise to a clean sample $\bm{x}^\text{(0)}$ from a data distribution $q(\bm{x}^\text{(0)})$ and turns it into a Gaussian noise $\bm{x}^{(T)}$, which is defined as 
\begin{equation}
q(\bm{x}^{(0)} , ...,\bm{x}^{(T)})=\prod\limits_{t = 1}^{T}{q{( \bm{x}^{(t)} | \bm{x}^{(t - 1)} )}} ,
\label{eq1}
\end{equation}
where ${q( \bm{x}^{(t)} | \bm{x}^{(t - 1)} )}=\mathcal{N}(\bm{x}^{(t)};\sqrt{1 - \beta_{t}}\bm{x}^{(t - 1)},\beta_{t}\bm{I})$ is the Markov diffusion kernel, $t=1, ..., T$, $T$ is the number of diffusion steps, and $\beta_{t}$ is a variance schedule. We have ${q( \bm{x}^{(t)} | \bm{x}^{(0)} )}=\mathcal{N}(\bm{x}^{(t)};\sqrt{{\overline{\alpha}}_{t}}\bm{x}^{(0)},(1-{\overline{\alpha}}_{t})\bm{I})$ by reparameterization with ${\alpha}_{t}=1-\beta_{t}$ , $\overline{\alpha}_t=\prod_{s = 0}^{t}\alpha_{s}$. $\bm{x}^{(t)}$ can be sampled by
\begin{equation}
\bm{x}^{(t)}=\sqrt{{\overline{\alpha}}_{t}}\bm{x}^{(0)}+\epsilon\sqrt{(1-{\overline{\alpha}}_{t})} ,
\label{eq2}
\end{equation}
where $\bm{\epsilon}$ is a standard Gaussian noise and $\epsilon\sim\mathcal{N}(0,\bm{I})$. When $T$ is large enough, $\bm{x}^{(T)}$ will eventually become a Gaussian noise.

The reverse process is also a Markovian process that denoises over a series of steps to generate meaningful data from the target distribution $q(\bm{x}^{(0)})$. 
The inverse process denoises the noise $\bm{x}^{(T)}$ from a distribution $p(\bm{x}^{(T)})$, which is defined as
\begin{equation}
p_\theta(\bm{x}^{(0)}, ...,\bm{x}^{(T-1)} | \bm{x}^{(T)}, \bm{c})=\prod\limits_{t = 1}^{T}{p_\theta{( \bm{x}^{(t-1)} | \bm{x}^{(t)}, \bm{c})}} ,
\label{eq3}
\end{equation}
where $p_\theta{( \bm{x}^{(t-1)} | \bm{x}^{(t)}), \bm{c}}=\mathcal{N}(\bm{x}^{(t-1)};\mu_\theta(\bm{x}^{(t)},t, \bm{c}),\sigma_t^2\bm{I})$, the mean $\bm{\mu}_\theta(\bm{x}^{(t)},t, \bm{c})$ is estimated by a neural network parameterized by $\bm{\theta}$, $\bm{c}$ is the latent condition encoding, and $\sigma_t^2$ is a step-dependent variance. 
$\bm{\mu}_\theta$ can be reparameterized as 
\begin{equation}
\mu_\theta(\bm{x}^{(t)},t, \bm{c})=\frac{1}{\sqrt{\alpha_{t}}}(\bm{x}^{(t)}-\frac{\beta_t}{\sqrt{1-\overline{\alpha}_{t}}} \bm{\epsilon}_{\bm{\theta}}(\bm{x}^{(t)},t, \bm{c})) ,
\label{eq4}
\end{equation}
where $\bm{\epsilon}_{\bm{\theta}}(\bm{x}^{(t)},t, \bm{c})$ is a neural network utilized to denoise the Gaussian noise from $\bm{x}^{(T)}$. 

The training objective is minimized by training $\bm{\epsilon}_{\bm{\theta}}(\bm{x}^{(t)},t, \bm{c})$ to approximate $\bm{\epsilon}$. The training objective is defined as
\begin{equation}
\mathcal{L}=\mathbb{E}_{t\sim[1:T],\bm{x^{(0)}}\sim q(\bm{x^{(0)}}),\epsilon\sim\mathcal{N}(0,\bm{I})}\parallel \bm{\epsilon}-\bm{\epsilon}_{\bm{\theta}}(\sqrt{\overline{\alpha}_{t}}\bm{x}^{(0)}+\sqrt{1-\overline{\alpha}_{t}}\epsilon,t,\bm{c})\parallel ,
\label{eq5}
\end{equation}
where $t$ is sampled from the uniform distribution over {1,2, ..., $T$}, $q(\bm{x^{(0)}})$ is the distribution of $\bm{x^{(0)}}$, and $\bm{\epsilon}$ is the Gaussian noise.

\subsection{Diffusion-based 3D anomaly reconstruction}
We formulate the point cloud reconstruction task of the anomaly-free model as the conditional generation, which decodes the explicit displacement with the target distribution $q(\mathcal{P}_\text{r}|\mathcal{P}_\text{a}^{(T)}, \bm{c})$, where $\bm{c}$ is the decoding condition. 
The essential question of anomaly detection in this paper is how to conditional reconstruct anomaly-free shapes on the reference of input point clouds with different spatial transformations. 
Since there is a high similarity of global features between abnormal and normal samples during self-supervised reconstruction, the most immediate approach is to extract an efficient global feature from input to serve as an auxiliary conditional embedding for the denoising function $\bm{\epsilon}_{\bm{\theta}}$. 
We implement the encoding of latent shape embedding $\bm{c}$ as a conditional input to guide reconstruction in the reverse diffusion process.

\subsubsection{Latent shape embedding}
The feature encoder aims to encode the point cloud to the latent shape embedding $\bm{c}$ with high-level features for the conditional generation process. 
Different from other global-local extracting methods~\cite{dgcnn, pointbert}, we focus more on extracting global features, which characterize the semantic information of shape and pose of most anomaly-free regions in the point cloud. 
The feature encoder mainly consists of cascaded multi-layer perceptions (MLP) based on PointNet\cite{PointNet}. 
It implements max-pooling after mapping $\mathcal{P}_\text{a}^{(0)}$ to different dimensions and then compresses them to extract the global shape embedding. 


\subsubsection{Step-wise displacement decoding}
To achieve point cloud reconstruction with transformation consistency while preserving the structure of non-anomalous regions, our method injects latent shape embedding $\bm{c}$ to the decoder at each step of the reverse diffusion process, as shown in Fig.~\ref{fig:framework}. 
In principle, in the training phase, $\bm{\epsilon}_{\bm{\theta}}$ learns the added Gaussian noise in the forward diffusion process by the decoder to model the conditional probability distribution. 
Conditionally generating target shapes from $N \times 3$ Gaussian noise is a straightforward approach, but it is afflicted by the issues of reconstructing the point cloud details and transform consistency. Learning the relative deformation of points for anomalous objects is more efficient.
Considering the mapping degradation of the vanilla autoencoder in the reconstruction training phase~\cite{denoisead}, we utilize the Gaussian noise of the forward process Eq.~\ref{eq2} to fully mask the point cloud object directly without blind spots, preventing the decoding process from receiving negative state shapes. 
The masked points $\mathcal{P}_\text{a}^{(T)}$ and latent shape embedding $\bm{c} \in \mathbb{R}^{256}$ are as the inputs of the SWD decoder. The point-wise displacement vector $\mathrm{\Delta} ^{(t)}$ is generated at each step of the iterative process thus disentangling the prediction noise and the desired anomaly-free shape. 
The reverse process can be defined according to Eq.~\ref{eq3} and the displacement vector $\mathrm{\Delta} ^{(t)}$ can be represented by 
\begin{equation}
\mathrm{\Delta}^{(t-1)}=\frac{1}{\sqrt{\alpha_{t}}}(\mathrm{\Delta}^{(t)}-\frac{\beta_t}{\sqrt{1-\overline{\alpha}_{t}}} \bm{\epsilon}_{\bm{\theta}}(\mathrm{\Delta}^{(t)},\beta_t, \bm{c}))+\sigma\bm{\epsilon} ,
\label{eq6}
\end{equation}
where $\sigma$ is the variance. 
A PointwiseNet is adopted for $\bm{\epsilon}_{\bm{\theta}}$ to decode the $\mathrm{\Delta} ^{(t-1)}$ from the previous step and $\bm{c}$. $\beta_t$ is used to generate trigonometric position embedding $\bm{e}_\text{p}$ = ($\beta_t$, sin($\beta_t$), cos($\beta_t$)). 
$\bm{e}_\text{p}$ is concatenated with $\bm{c}$ and then fed into the concatenate-squash linear module of PointwiseNet with a residual function. 
The output reconstructed point cloud at the $t$ step is $\mathcal{P}_\text{r}^{(t)}=\mathcal{P}_\text{r}^{(t+1)}+\mathrm{\Delta}^{(t)}$.
The registered original and reconstructed objects are distinguished from the anomalous shape by the anomaly scores based on the conditioned distance function.

\subsection{3D anomaly simulation strategy}
Given that a small number of normal samples is not conducive for the model to learning diverse and essential features, we propose the Patch-Gen strategy to simulate the defects from anomaly-free shapes for training data augmentation. 
Patch-Gen encourages the reconstruction model to learn to detect irregularity, where the anomaly-free point clouds and their diverse anomaly patterns are integrated into training pairs and are utilized to learn the discrimination feature between normal and anomalous surfaces. 
The intuition is that the diversity of simulated negative samples forces our network to learn how to reconstruct anomaly-free shapes instead of memorizing their complete outfits.

\begin{figure}[t]
\centering
\includegraphics[width=0.8\linewidth]{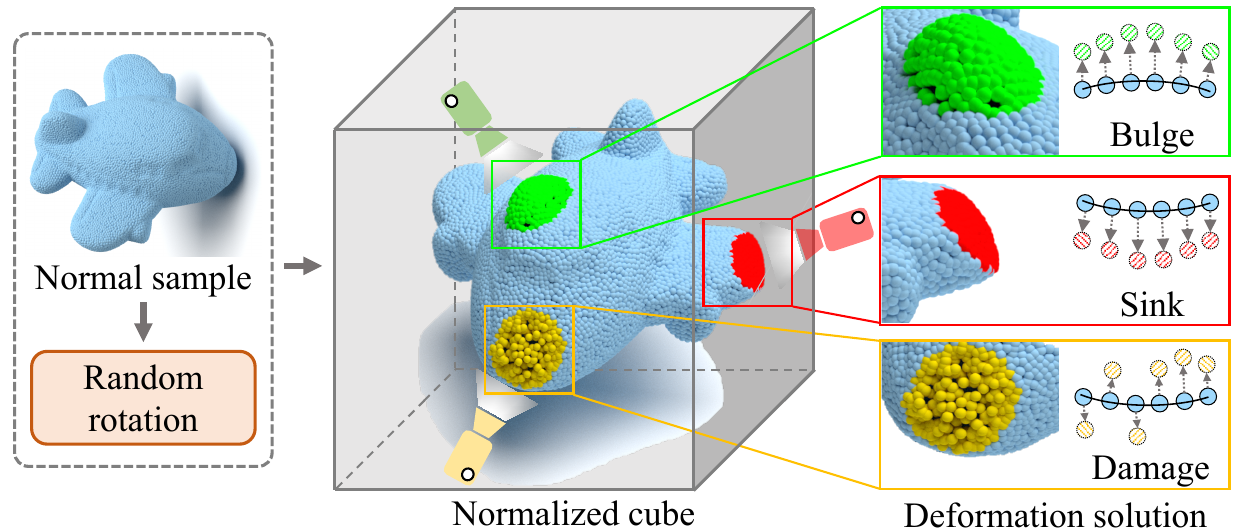}
\caption{Illustration of Patch-Gen, the 3D anomaly simulation strategy. 
The input normal point cloud is first randomly rotated.
On the surface of the normalized cube, we randomly select viewpoints to find the nearest patch of points.
The selected points are then transformed into irregular defects according to the specific deformation solution.}
\label{fig:patch}
\end{figure}

As shown in Fig.~\ref{fig:patch}, the input normal point cloud is first randomly rotated.
The random spatial rotation is designed to improve the generalization capability for test samples with very different spatial transformations, as defined by:
\begin{equation}
    \mathcal{P}_\text{a}=\mathcal{P} \cdot \mathcal{R}, 
\end{equation}
where $\mathcal{P}$ is the normal sample and $\mathcal{R} \in \mathbb{R}^{3 \times 3}$ is obtained by randomly selecting rotation angles for all three axes.
In addition to global shape awareness of the model by the random rotation, we further perform a fine granularity of the anomaly simulation.
We randomly take a viewpoint $\mathcal{P}_v$ from the surface of the cube.
Therefore, the patch of nearest $N$ points $\mathcal{P}_n$ from $\mathcal{P}_a$ can be determined according to the $\mathcal{P}_v$. 
The shape augmentation scheme Patch-Gen is defined as follows:
\begin{equation}
\mathcal{P}_n = \mathcal{P}_n + S \cdot \textit{normalize}(\mathcal{P}_n - \mathcal{P}_v ) \odot \mathcal{T},
\label{eq7}
\end{equation}
where $nomalize$ represents a normalization operation on a vector, $S$ is a predefined hyper-parameter that controls the scaling of the patch points, and $\mathcal{T}$ is the translation matrix originating from a Gaussian distribution.
The $\mathcal{P}_\text{a}$ is finally obtained by only updating the patch region $\mathcal{P}_n$ while keeping the rest points unchanged.

With the proposed Patch-Gen, we can simulate the generation of multiple anomalies, which is mainly done by controlling $\mathcal{T}$. 
Bulge or sink can be generated by sorting $\mathcal{T}$ after sampling from the distribution, while damage can be generated by direct overlaying without manipulation.
Fig.~\ref{fig:vis_patch} further illustrates the contrast between the generated 
anomalies and actual ones, affirming that our approach can remarkably emulate real-world scenarios with a high degree of fidelity.

\subsection{Training objective}
In the reconstruction task of the object with $N$ points, the network learns a diffusion model with an $\mathbb{R}^{N \times 3} \rightarrow \mathbb{R}^{N \times 3}$ mapping relation. 
Iterative denoising under the semantic condition of point embedding realizes the prediction of point offsets. 
Concretely, the network is trained to learn the noise that needs to be eliminated to recover the anomaly-free shape with the $L_2$ distance between the ground truth and the denoised reconstructed points. 
We make use of the mean squared error (MSE) loss as the primary reconstruction loss which evaluates the mean squared error of the element-wise distances between $\mathcal{P}_\text{a}^{(0)}$ and $\mathcal{P}_\text{r}^{(0)}$. 
The MSE training loss is formulated as:
\begin{equation}
\mathcal{L}_{{\mathcal{P}}_\text{a}, {\mathcal{P}}_\text{r}}=\frac{1}{N}{\sum\limits_{i = 1}^{N}{_{p_\text{a} \in {\mathcal{P}}_\text{a}  ,p_\text{r} \in {\mathcal{P}}_\text{r}}{\parallel p_\text{a}  - p_\text{r}\parallel^2}}} .
\label{eq8}
\end{equation}

\section{Experiments}
\subsection{Datasets}

\begin{table}[t]
\centering
\resizebox{1.0\textwidth}{!}{
\begin{tabular}{lccccccccc}
\toprule
\textbf{Method} & \multicolumn{2}{c}{BTF\cite{btf}} & M3DM\cite{m3dm} & \multicolumn{2}{c}{PatchCore\cite{patchcore}}  & CPMF\cite{cpmf} & Reg3D-AD\cite{real3dad} & IMRNet\cite{shapenetad} & \textbf{Ours}\\
\cmidrule(r){2-3} \cmidrule(r){5-6}
\textbf{Feat.} & \textit{Raw} & \textit{FPFH} &\textit{PointMAE} & \textit{FPFH} & \textit{PointMAE} &\textit{ResNet} &\textit{PointMAE} &\textit{PointMAE} &\textit{Raw} \\
\midrule
Airplane & 0.730 & 0.520 & 0.434 & \textbf{0.882} & 0.726 & 0.701 & 0.716 & 0.762 & \underline{0.772} \\ 
Candybar & 0.539 & 0.630 & 0.552 & 0.541 & 0.663 & 0.552 & 0.685 & \textbf{0.755} &  \underline{0.696}\\ 
Car & 0.647 & 0.560 & 0.541 & 0.590 & 0.498 & 0.551 & 0.697 & \underline{0.711} & \textbf{0.713} \\ 
Chicken & 0.789 & 0.432 & 0.683 & \underline{0.837} & 0.827 & 0.504 & \textbf{0.852} & 0.780 & 0.714 \\ 
Diamond & 0.707 & 0.545 & 0.602 & 0.574 & 0.783 & 0.523 & \underline{0.900} & \textbf{0.905} & 0.685 \\ 
Duck & 0.691 & \underline{0.784} & 0.433 & 0.546 & 0.489 & 0.582 & 0.584 & 0.517 & \textbf{0.909} \\ 
Fish & 0.602 & 0.549 & 0.540 & 0.675 & 0.630 & 0.558 & \textbf{0.915} & \underline{0.880} & 0.692 \\ 
Gemstone & \textbf{0.686} & 0.648 & 0.644 & 0.370 & 0.374 & 0.589 & 0.417 & \underline{0.674} & 0.665 \\ 
Seahorse & 0.596 & \textbf{0.779} & 0.495 & 0.505 & 0.539 & \underline{0.729} & 0.762 & 0.604 & 0.720 \\ 
Shell & 0.396 & \underline{0.754} & 0.694 & 0.589 & 0.501 & 0.653 & 0.583 & 0.665 & \textbf{0.840} \\
Starfish & 0.530 & 0.575 & 0.551 & 0.441 & 0.519 & \underline{0.700} & 0.506 & 0.674 & \textbf{0.701} \\ 
Toffees & 0.703 & 0.462 & 0.450 & 0.565 & 0.585 & 0.390 & \textbf{0.827} & \underline{0.774} & 0.703 \\ 
\midrule
\textbf{Average} & 0.635 & 0.603 & 0.552 & 0.593 & 0.595 & 0.586 & 0.704 & \underline{0.725} & \textbf{0.734} \\
\bottomrule
\end{tabular}
}
\caption{Image-level anomaly detection AUROC on Real3D-AD dataset. 
We highlight the best result in \textbf{bold} and the second best result with an \underline{underline}.}
\label{tab:real3dad}
\end{table}

\subsubsection{Real3D-AD}~\cite{real3dad} is a 3D anomaly detection dataset based on real samples, exhibiting a higher point precision and spatial distance per point cloud. 
Each category contains 4 training samples and 100 test samples. 
The training set contains 360° complete surface point clouds of the objects, which are obtained by manually calibrating and stitching the scans of multiple sides of the objects. 
The test samples are scans only one side with a huge difference from the training set. 
The distribution of the point clouds also varies among the total 12 categories, further deepening the detection difficulty compared to 2D scenes.

\subsubsection{Anomaly-ShapeNet}~\cite{shapenetad} is a 3D anomaly detection, crafted through modifications to the synthetic samples found in ShapeNetCorev2~\cite{shapenet}. 
It contains 40 diverse categories, featuring over 1600 samples of its complete surface point clouds.
Each category's training set contains merely 4 samples, while the test sets are designed to assess the model's performance across both normal and a spectrum of abnormal samples.
It widely increases the anomaly types while keeping the number of points the same as the previous studies, which places higher demands on the robustness and generality of the proposed algorithms.

\subsection{Evaluation metrics}
For image-level anomaly detection, the Area Under the Receiver Operating Curve (AUROC) is utilized in line with established practices.
For the evaluation of pixel-level anomalies, the AUROC metric is similarly applied in the context of point segmentation accuracy.
A value of 0.5 of the AUROC score denotes no discriminative capability (equivalent to random guessing), whereas a score of 1.0 indicates perfect discrimination between positive and negative classes.

\subsection{Implementation details}
Our methodology is implemented using PyTorch~\cite{pytorch} with end-to-end training across the network.
The optimization is performed using the Adam optimizer, starting at an initial learning of 0.001.
The training process involves a total batch size of 128 across 40,000 iterations for comprehensive learning.
All input point clouds undergo a preprocessing step where they are randomly downsampled to a fixed size of 4096 and 2048 points on Real3D-AD and Anomaly-ShapeNet, respectively. 
Additionally, we normalized these point clouds by setting their center of gravity as the origin of coordinates and scaling their dimensions to fall within the range of -1 to 1, optimizing for the diffusion process.

\begin{table}[t]
\centering
\resizebox{1.0\textwidth}{!}{
\begin{tabular}{lccccccccc}
\toprule
\textbf{Method} & \multicolumn{2}{c}{BTF\cite{btf}} & M3DM\cite{m3dm} & \multicolumn{2}{c}{PatchCore\cite{patchcore}}  & CPMF\cite{cpmf} & Reg3D-AD\cite{real3dad} & IMRNet\cite{shapenetad} & \textbf{Ours}\\
\cmidrule(r){2-3} \cmidrule(r){5-6}
\textbf{Feat.} & \textit{Raw} & \textit{FPFH} &\textit{PointMAE} & \textit{FPFH} & \textit{PointMAE} &\textit{ResNet} &\textit{PointMAE} &\textit{PointMAE} &\textit{Raw} \\
\midrule
Ashtray & 0.578 & 0.420 & 0.577 & 0.587 & 0.591 & 0.353 & 0.597 & \underline{0.671}     &\textbf{0.833} \\ 
Bag & 0.410 & 0.546 & 0.537 & 0.571 & 0.601 & 0.643 & \underline{0.706} & 0.660         &\textbf{0.719}	\\ 
Bottle & 0.558 & 0.404 & 0.584 & 0.614 & 0.588 & 0.469 & 0.569 &\underline{0.631}      &\textbf{0.750}	\\ 
Bowl & 0.470 & 0.581 & 0.579 & 0.558 & 0.547 & \underline{0.679} & 0.548 & 0.676        &\textbf{0.751}	\\ 
Bucket & 0.469 & 0.517 & 0.405 & 0.510 & 0.577 & 0.542 & \underline{0.681} & 0.676      &\textbf{0.719}	\\ 
Cap & 0.509 & 0.562 & 0.599 & 0.645 & 0.583 & 0.601 & 0.632 & \underline{0.704}         &\textbf{0.726}	\\ 
Cup & 0.462 & 0.598 & 0.548 & 0.593 & 0.583 & 0.498 & 0.524 & \underline{0.700}         &\textbf{0.767}	\\ 
Eraser & 0.525 & \underline{0.719} & 0.627 & 0.657 & 0.677 & 0.689 & 0.343 & 0.548      &\textbf{0.890}	\\ 
Headset & 0.447 & 0.505 & 0.597 & 0.610 & 0.609 & 0.551 & 0.574 & \underline{0.698}     &\textbf{0.767}	\\ 
Helmet & 0.508 & 0.569 & 0.488 & 0.465 & 0.495 & 0.532 & 0.491 &\underline{0.603}      &\textbf{0.704}	\\ 
Jar & 0.420 & 0.424 & 0.441 & 0.472 & 0.483 & 0.610 & 0.592 & \underline{0.780}         &\textbf{0.838}	\\ 
Microphone & 0.563 & 0.671 & 0.357 & 0.388 & 0.488 & 0.509 & 0.414 & \underline{0.755}  &\textbf{0.762}	\\ 
Shelf & 0.164 & 0.609 & 0.564 & 0.494 & 0.523 & 0.685 & \underline{0.688} & 0.603       &\textbf{0.696}	\\ 
Tap & 0.549 & 0.553 & 0.747 & \underline{0.760} & 0.498 & 0.528 & 0.659 & 0.686         &\textbf{0.818}	\\ 
Vase & 0.517  & 0.464  & 0.534  & 0.554  & 0.582  & 0.514  & 0.576 & \underline{0.629}  &\textbf{0.734}\\ 
\midrule
\textbf{Average} & 0.493 & 0.528 & 0.552 & 0.568 & 0.562 & 0.559 & 0.572 & \underline{0.659} &\textbf{0.749} \\
\bottomrule
\end{tabular}
}
\caption{Image-level anomaly detection AUROC on Anomaly-ShapeNet dataset.
We highlight the best result in \textbf{bold} and the second best result with an \underline{underline}.}
\label{tab:shapenetad}
\end{table}

\subsection{Main results}
We conduct experiments on Real3D-AD~\cite{real3dad} based on real sampling and Anomaly-ShapeNet~\cite{shapenetad} based on simulation.

As shown in Table~\ref{tab:real3dad}, we first compare the image-level AUROC metric with current cutting-edge 3D anomaly detection models on Real3D-AD.
It shows that our method achieves the best performance using only raw point cloud data, while most of the existing methods use Fast Point Feature Histograms (FPFH) operator~\cite{fpfh} or ShapeNet~\cite{shapenet} pre-trained PointMAE~\cite{pointmae} as feature extractor.
Due to significant disparities in quantity, size, and distribution among different categories of point clouds in Real3D-AD, scoring variations across categories are more pronounced with other methods. 
For instance, numerous methods perform under 0.5 in certain categories, indicating their inadequacy in extracting meaningful features while facing challenging samples.
In contrast, our method not only exhibits superior performance in 3D anomaly detection across the majority of categories but also achieves the best overall average across all categories. This demonstrates the strong generalizability and robustness of our approach.

We further evaluate our method on Anomaly-ShapeNet in Table~\ref{tab:shapenetad}, which encompasses a broader array of categories and a greater diversity of defect types.
Compared to Real3D-AD, Anomaly-ShapeNet significantly enhances the diversity of defects, wherein the increased variety of defect types further escalates the complexity of detection tasks. 
The results highlight the exceptional performance of our method across all evaluated categories, demonstrating an average improvement of 9\% on AUROC relative to the approaches previously utilized.

\begin{table*}[t]
\centering
\resizebox{0.8\textwidth}{!}{
\begin{tabular}{ccccc|cc}
\toprule
Model  &Diffusion  & Condition &Relative &Patch-Gen  &I-AUROC &P-AUROC \\
\midrule
A  &\Checkmark  &\XSolidBrush   &\XSolidBrush   &\XSolidBrush   &0.586 &0.524 \\
B  &\Checkmark  &\Checkmark     &\XSolidBrush   &\XSolidBrush   &0.667 &0.513 \\
C  &\Checkmark  &\Checkmark     &\Checkmark     &\XSolidBrush   &0.712 &0.573 \\
D  &\Checkmark  &\Checkmark     &\Checkmark     &\Checkmark     &\textbf{0.734} &\textbf{0.592} \\
\bottomrule
\end{tabular}
}
\caption{Ablation studies for 3D adaptation components on Real3D-AD dataset.
}
\label{tab:ablation}
\end{table*}

\subsection{Ablation study}
To delve into the effect of individual components, we conduct ablation experiments on the Real3D-AD dataset.
To fully demonstrate and compare the performance of the models, we report both image-level and pixel-level results with I-AUROC and P-AUROC, respectively.

\subsubsection{Main component}
Table~\ref{tab:ablation} compares the performance of different variants from R3D-AD, which includes the influence of the denoising condition embedding, displacement-based reconstruction way, and the data augmentation strategy of Patch-Gen. 
Model A is denoted as our baseline, which is a vanilla DDPM model for point cloud reconstruction. 
Introducing a condition into the DDPM (Model B) significantly boosts performance, particularly in terms of I-AUROC, which sees a 13.8\% increase to 0.667. 
Model C, which predicts point displacements based on conditional DDPM, preserving detailed structural information while accommodating the relative displacement of points contributes to a notable 6.0\% gain in P-AUROC over Model B.
Model D is trained under the conditions of shape embedding with the Patch-Gen strategy.
Considering that the defective portion contains only a small portion of the original point cloud, we try to reconstruct the relative displacement in a way that preserves as much detail as possible,  which is effective for both 3D anomaly detection and segmentation.

\subsubsection{Patch-Gen}
Table.~\ref{tab:patch} analyzes the influence of two key parameters in Patch-Gen: the selection points ratio and the scaling points factor.

The \textbf{selection points ratio} from Table.~\ref{subtab:select} determines the proportion of points in the point cloud that are selected for transformation. 
Our findings suggest that a selection ratio of 1/32 achieves the best performance. 
It appears that this ratio provides a balanced trade-off between maintaining sufficient structure for anomaly detection and introducing enough variation to simulate anomalies effectively. 
Notably, as the ratio increases beyond 1/16, both I-AUROC and P-AUROC scores decrease in severity, since real defects only account for a small portion of the overall point cloud, a wide selection of points not only destroys the structure of the original point cloud, but also makes the distribution of the training and test sets inconsistent.

The \textbf{scaling points factor} is the intensity of the random transformation applied to the selected points, as detailed in Table~\ref{subtab:scale}. 
The optimal performance is observed at a scaling factor of 0.1, which implies that minor transformations are more effective for simulating anomalies without significantly altering the original data distribution. 
Larger scaling factors lead to a consistent decline in performance, underscoring the importance of subtle transformations for preserving the utility of the simulated anomalies for detection tasks.

\begin{table}[t]
\centering
\subfloat[\textbf{Selection points ratio.}\label{subtab:select}]{
    \begin{tabular}{r|cc}
    \toprule
    ratio & I-AUROC & P-AUROC \\
    \midrule
    1/64 &0.716 &0.584 \\
    \rowcolor{lightgray} 1/32 &\textbf{0.734} &\textbf{0.592} \\
    1/16 &0.727 &0.579 \\
    1/8  &0.683 &0.528 \\
    \bottomrule
    \end{tabular}
}
\subfloat[\textbf{Scaling points factor.}\label{subtab:scale}]{
    \begin{tabular}{r|cc}
    \toprule
    factor & I-AUROC & P-AUROC \\
    \midrule
    \rowcolor{lightgray} 0.1 &\textbf{0.734} &\textbf{0.592} \\
    0.2 &0.727 &0.572 \\
    0.4 &0.715 &0.554 \\
    0.8 &0.661 &0.517 \\
    \bottomrule
    \end{tabular}
}
\caption{Ablation studies for Patch-Gen implementation on Real3D-AD dataset.
Default settings are marked in \colorbox{lightgray}{gray}.}
\label{tab:patch}
\end{table}

\begin{figure}[t]
\centering
\includegraphics[width=\linewidth]{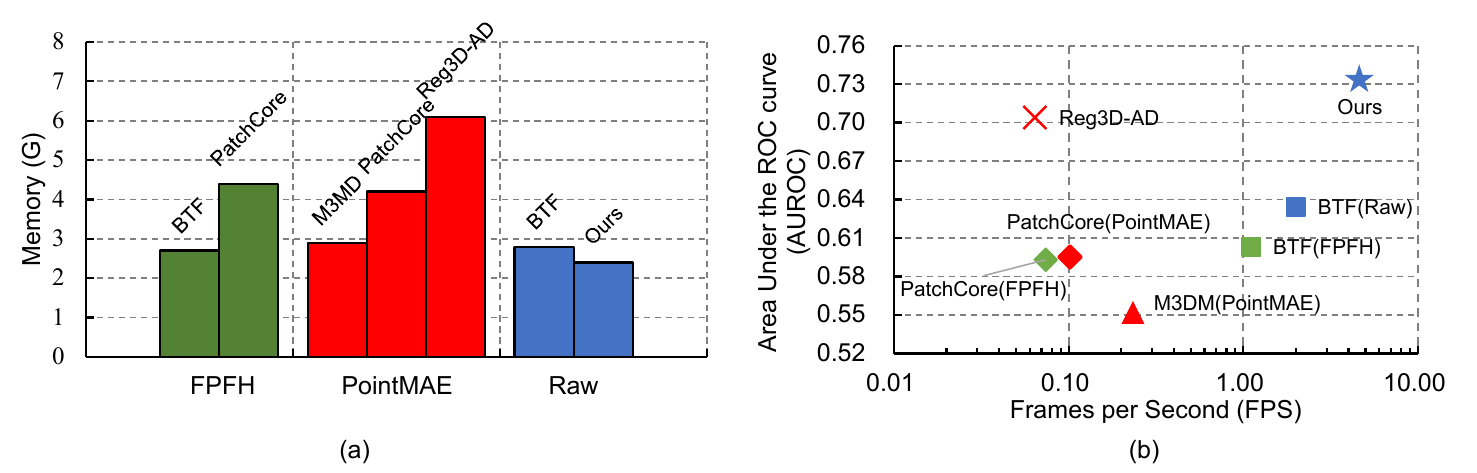}
\caption{Memory and time cost during inference on Real3D-AD dataset.
\textbf{(a)} Memory usage comparison between different models.
\textbf{(b)} 3D anomaly detection performance vs. frames per second on an NVIDIA RTX 3090 GPU. 
Our R3D-AD outperforms all previous methods on both accuracy and efficiency by a significant margin.}
\label{fig:cost}
\end{figure}

\subsubsection{Memory and time cost}
As depicted in Figure~\ref{fig:cost}, we evaluate the disparity in both storage consumption and inference time of our model under identical experimental conditions, compared to existing methods.
Regarding memory usage, our approach demonstrates a marked superiority by employing raw coordinate features instead of FPFH or PointMAE features, significantly reducing the memory footprint. 
Since no memory bank exists, our method is also more space-efficient compared to BTF which also uses raw features.
Moreover, our method eliminates the necessity to compare all the features in memory, substantially increasing operational efficiency. 
The implementation of Patch-Gen inherently bestows our model with exceptional robustness, enabling precise reconstruction of point clouds from various angles without the need for the time-intensive RANSAC alignment process required by Reg3D-AD.

\begin{figure}[t]
\centering
\includegraphics[width=\linewidth]{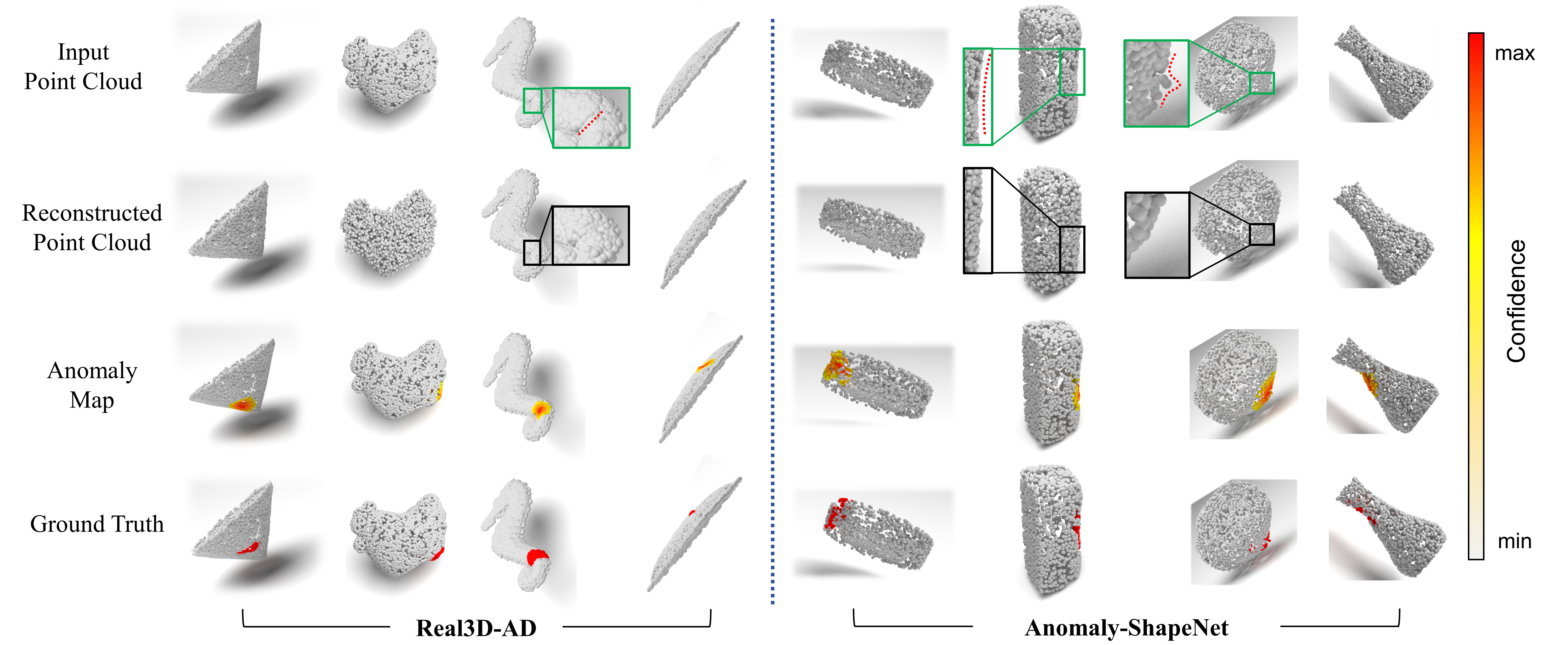}
\caption{Qualitative analysis on Real3D-AD dataset and Anomaly-ShapeNet dataset.
The anomaly map is obtained directly by calculating the differences between the input and reconstructed point clouds, where deeper colors represent more confidence.
}
\label{fig:vis}
\end{figure}

\subsection{Qualitative results}
Figure~\ref{fig:vis} presents some qualitative outcomes, with varying shades of color indicating different levels of anomaly scores.
We select several representative defective samples to demonstrate the robustness of our algorithm. 
The left four columns display samples from Real3D-AD, while the right four columns samples are from Anomaly-ShapeNet. 
The illustration reveals that our R3D-AD algorithm has precisely reconstructed the defective portions of the point cloud across various samples: the deep \textit{sink} in the Seahorse sample, the \textit{concavity} in the Bag sample, and the \textit{bulge} in the Jar sample. 
Leveraging the accurately reconstructed point clouds, final point cloud segmentation maps are also produced, further evidencing the efficacy of our approach.

\section{Conclusion}
In this work, we presented R3D-AD, a novel reconstructive 3D anomaly detection model based on conditional diffusion.
Our goal is to overcome the limitations faced by current 3D anomaly detection methods, such as the inefficiencies due to the memory bank module and low performance caused by incorrect rebuilds with MAE.
To address these challenges, we leverage the diffusion process for full reconstruction, followed by a direct comparison between the input and the reconstructed point cloud to obtain the final anomaly score.
The embedded latent variable that spans the decoding process, step-wisely generating point-level displacements from the noise to the target anomaly-free sample.
We also propose Patch-Gen, a data augmentation tailored for point cloud anomaly simulation.
Extensive experiments conducted on 3D anomaly benchmarks validate the superiority of our R3D-AD in comparison to state-of-the-art alternatives in terms of both accuracy and versatility.
\section*{Acknowledgements}
This work was supported in part by the Pioneer and Leading Goose R\&D Program of Zhejiang (Grant No. 2022C01051), in part by the National Natural Science Foundation of China (Grant No. 52375271, 52275274), and in part by the Natural Science Foundation of Zhejiang Province (Grant No. LY23E050011).

\bibliographystyle{splncs04}
\bibliography{main}

\appendix
\section{Appendix}

\subsection{Additional implement details}

\subsubsection{Patch-Gen pseudocode}
We formulate the process of the proposed 3D anomaly simulation strategy Patch-Gen in Algorithm~\ref{algo:patch-gen}.
The procedure begins by taking an initial point cloud $\mathcal{P}$
as input and aims to produce an augmented point cloud $\mathcal{P}_a$ that reflects the addition of anomaly.
The rotation matrix $\mathcal{R}$ is obtained by applying arbitrary rotation angles to all the rotation axes.
The translation matrix $\mathcal{T}$ is sampled from a Gaussian distribution, and after normalization and scaling, it dictates the displacement of the nearest points towards the viewpoint, while the rest of the point cloud remains unchanged. 
The anomaly point cloud representing damage types can be acquired through direct manipulation of matrix $\mathcal{T}$, derived from random sampling procedures. 
By sorting the matrix $\mathcal{T}$, we can further simulate defects such as bulge and sink. 
This perspective can be likened to a gravitational force acting as an anchor, exerting influence on the patch points within the domain.

\begin{algorithm}[h]
\caption{Patch-Gen} \label{algo:patch-gen}
\begin{algorithmic}
\Require {
$\mathcal{P}$: input point cloud \\
$N$: number of points to select \\
$S$: scaling factor for transformation
}
\Ensure $P_a$: augmented point cloud
\State $\mathcal{R} \gets$ random rotation matrix \Comment $\mathbb{R}^{3 \times 3}$
\State $\mathcal{P}_a = \mathcal{P} \cdot \mathcal{R} $ \Comment apply rotation
\State $\mathcal{P}_v \gets$ random viewpoint \Comment $\mathbb{R}^{1 \times 3}$
\State $\mathcal{P}_n = \textit{NN}(\mathcal{P}_a, \mathcal{P}_v, N)$ \Comment select $N$ nearest neighbor points to the viewpoint
\State $\mathcal{T} \gets$ random translation matrix \Comment $\mathbb{R}^{N \times 3}$
\State $\mathcal{P}_n = \mathcal{P}_n + S \cdot \textit{normalize}(\mathcal{P}_n - \mathcal{P}_v ) \odot \mathcal{T}$ \Comment update selected points only
\end{algorithmic}
\end{algorithm}

\subsubsection{R3D-AD pseudocode}
To further clarify the overall architecture of the proposed network R3D-AD, we provide the training and testing iteration procedures more compactly in Algorithm~\ref{algo:training} and Algorithm~\ref{algo:testing}, respectively.

During \textbf{training}, anomalies are simulated by Patch-Gen, and noise is artificially added following a Gaussian distribution. 
The model predicts this noise and calculates a displacement to correct for it. 
The reconstruction loss is measured by comparing the original and corrected point clouds.

\begin{algorithm}[h]
\caption{R3D-AD training iteration} \label{algo:training}
\begin{algorithmic}
\Require $\mathcal{P}$: input point cloud
\Ensure $\mathcal{L}$: reconstruction loss

\State $\mathcal{P}^{'} \sim \mathrm{Uniform}(normalize (\mathcal{P}))$ \Comment normalize and downsample the input point cloud
\State $\mathcal{P}_a^{(0)} = \textit{Patch-Gen}(\mathcal{P}^{'})$ \Comment 3D anomaly simulation strategy (Algorithm~\ref{algo:patch-gen})
\State $c = \textit{PointNet}(\mathcal{P}_a^{(0)})$ \Comment feature extraction

\State $\bp_a^{(0)} \sim q(\bp_a^{(0)})$ \Comment point distribution
\State $t \sim \mathrm{Uniform}(\{1, \dotsc, T\})$ \Comment step distribution
\State $\bepsilon\sim\mathcal{N}(\bzero,\bI)$ \Comment noise distribution

\State $\bmu = \bepsilon_\theta(\sqrt{\bar\alpha_t} \bp_a^{(0)} + \sqrt{1-\bar\alpha_t}\bepsilon, \textbf{\textit{c}}, t)$ \Comment noise prediction
\State $\Delta = \frac{1}{\sqrt{\bar\alpha_t}}\left( \sqrt{\bar\alpha_t} \bp_a^{(0)} + \sqrt{1-\bar\alpha_t} (\bepsilon - \bmu) \right)$ \Comment displacement prediction
\State $\mathcal{L} = \left\| \mathcal{P}^{'} - (\mathcal{P}_a^{(0)} + \Delta) \right\|^2$ \Comment relative reconstruction loss
\end{algorithmic}
\end{algorithm}

During \textbf{testing}, noise is progressively removed from a simulated noisy version of the cloud, aiming to reconstruct its anomaly-free outfits. 
The anomaly score is assessed by comparing the clusters after KNN of the original and reconstructed point clouds. 

\begin{algorithm}[h]
\caption{R3D-AD testing iteration} \label{algo:testing}
\begin{algorithmic}
\Require $\mathcal{P}$: input point cloud
\Ensure $\mathcal{A}$: anomaly score

\State $\mathcal{P}^{'} \sim \mathrm{Uniform}(normalize (\mathcal{P}))$ \Comment normalize and downsample the input point cloud
\State $c = \textit{PointNet}(\bp^{'})$ \Comment feature extraction

\State $\Delta^{(T)} \sim \mathcal{N}(\bzero, \bI)$
    \For{$t=T, \dotsc, 1$}
      \State $\bz \sim \mathcal{N}(\bzero, \bI)$ if $t > 1$, else $\bz = \bzero$
      \State $\mathrm{\Delta}^{(t-1)}=\frac{1}{\sqrt{\alpha_{t}}}(\mathrm{\Delta}^{(t)}-\frac{1-\alpha_t}{\sqrt{1-\overline{\alpha}_{t}}} \bm{\epsilon}_{\bm{\theta}}(\mathrm{\Delta}^{(t)},\bm{c}, t))+\sigma_t \bz$
    \EndFor
\State $\widehat{\mathcal{P}} = \mathcal{P}^{'} + \Delta^{(0)}$ \Comment reconstructed point cloud

\State $\widehat{cluster} = \textit{KNN}(\widehat{\mathcal{P}}, k)$ \Comment reconstructed point-cluster
\State $cluster = \textit{KNN}(\mathcal{P}^{'}, k)$ \Comment input point-cluster

\State $\mathcal{A} = \left\|cluster - \widehat{cluster} \right\|^2$ \Comment euclidean distance for point-cluster
\end{algorithmic}
\end{algorithm}

\begin{figure}[t]
\centering
\includegraphics[width=0.8\linewidth]{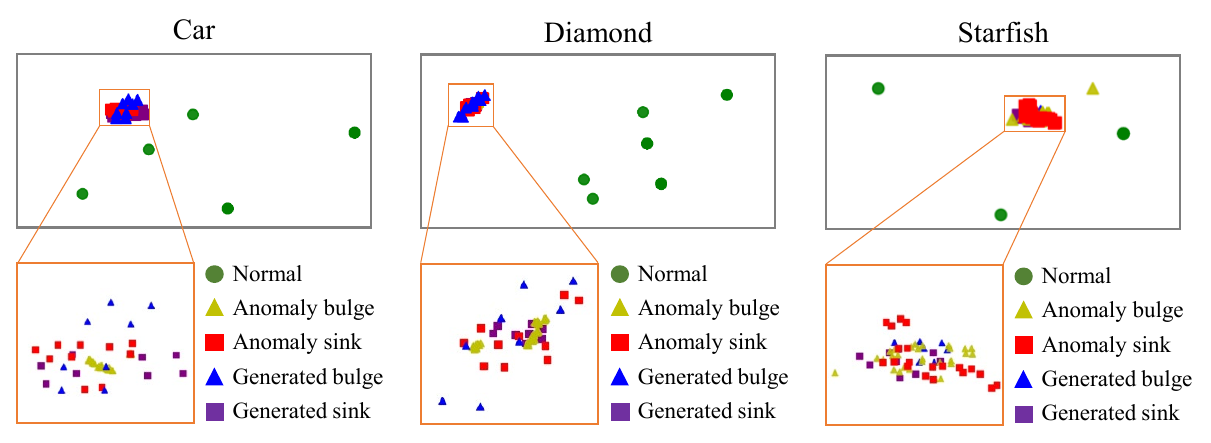}
\caption{t-SNE visualization on Real3D-AD.}
\label{tsne}
\end{figure}
\begin{table}[t]
\centering
\begin{tabular}{l|cc||c}
\toprule
Anomaly type & Bulge & Sink & Oracle \\
\midrule
Airplane & 1.31 & 1.35 & 1.58 \\
Candybar & 2.43 & 2.30 & 2.54 \\
Car & 1.15 & 1.23 & 1.37 \\
Chicken & 3.50 & 2.92 & 4.02 \\
Diamond & 0.84 & 0.83 & 0.97 \\
Duck & 1.53 & 1.29 & 1.67 \\
Fish & 1.42 & 1.45 & 1.57 \\
Gemstone & 2.58 & 5.23 & 5.26 \\
Seahorse & 2.37 & 2.35 & 2.45 \\
Shell & 1.30 & 1.29 & 1.40 \\
Starfish & 2.47 & 2.46 & 2.64 \\
Toffees & 1.73 & 1.71 & 1.79 \\ 
\bottomrule
\end{tabular}
\caption{PSNR of generated anomalous with Patch-Gen on Real3D-AD.}
\label{psnr}
\end{table}

\subsection{Additional experiments}

\subsubsection{Quality of the generated anomalies}
The proposed 3D anomaly simulation strategy Patch-Gen is designed to address the problem of the lack of 3D anomalous samples in the training phase.

T-distributed Stochastic Neighbor Embedding (t-SNE)~\cite{tsne} is particularly effective at visualizing high-dimensional samples by giving each data point a corresponding location in a low-dimensional map, allowing complex data to be understood at a glance. 
We follow~\cite{cutpaste} and use the t-SNE to validate the quality and effectiveness of our generated anomaly samples.
As shown in Fig~\ref{tsne}, the generated anomalies are clearly distinguished from normal samples and overlap with real anomalous samples, which strengthens our model to reconstruct well on unseen anomalies.

Peak Signal-to-Noise Ratio (PSNR) is an engineering term that quantifies the quality of the reconstruction of a signal.
PSNR is typically measured in decibels (dB) and calculated based on the mean squared error between the origin and the reconstruction.
The higher the PSNR value, the better the quality of the reconstruction. 
In Table~\ref{psnr}, the PSNR is computed by comparing the generated samples with real anomalies.
We randomly select two normal samples to calculate their PSNR, and we average the PSNR obtained from multiple times of randomization to obtain the upper bound of the PSNR limit for each category.
The Oracle PSNR servers are a reference to the generation quality.

\begin{table}[t]
\centering
\begin{tabular}{cccc|cc||c}
\toprule
\multicolumn{2}{c}{Training} &\multicolumn{2}{c|}{Testing} &\multirow{2}{*}{I-AUROC}  &\multirow{2}{*}{CD} &\multirow{2}{*}{Oracle}\\
Dataset &Category &Dataset &Category & & & \\
\midrule
    Real3D-AD &Airplane &ShapeNetCore.v2 &Airplane &- &0.032 &0.001\\
    Real3D-AD &Car &ShapeNetCore.v2 &Car &- &0.077 &0.004\\
\midrule
    ShapeNetCore.v2 &Airplane &Real3D-AD &Airplane &0.614 &- &0.772\\
    ShapeNetCore.v2 &Car &Real3D-AD &Car &0.601 &- &0.713\\
\midrule
    Anomaly-ShapeNet &\{bowl0..3\} &Anomaly-ShapeNet &bowl4 &0.715 &- &0.744 \\
\bottomrule
\end{tabular}
\caption{Generalization capability of R3D-AD for unseen data.}
\label{trans}
\end{table}
\begin{table}[h!]
\centering
\resizebox{1.0\textwidth}{!}{
\begin{tabular}{lccccccccc}
\toprule
\textbf{Method} & \multicolumn{2}{c}{BTF\cite{btf}} & M3DM\cite{m3dm} & \multicolumn{2}{c}{PatchCore\cite{patchcore}}  & CPMF\cite{cpmf} & Reg3D-AD\cite{real3dad} & IMRNet\cite{shapenetad} & \textbf{Ours}\\
\cmidrule(r){2-3} \cmidrule(r){5-6}
\textbf{Feat.} & \textit{Raw} & \textit{FPFH} &\textit{PointMAE} & \textit{FPFH} & \textit{PointMAE} &\textit{ResNet} &\textit{PointMAE} &\textit{PointMAE} &\textit{Raw} \\
\midrule
ashtray0 & 0.578 & 0.420 & 0.577 & 0.587 & 0.591 & 0.353 & 0.597 & 0.671 & \textbf{0.833} \\ 
bag0 & 0.410 & 0.546 & 0.537 & 0.571 & 0.601 & 0.643 & 0.706 & 0.660 & \textbf{0.720} \\ 
bottle0 & 0.597 & 0.344 & 0.574 & 0.604 & 0.513 & 0.520 & 0.486 & 0.552 & \textbf{0.733} \\ 
bottle1 & 0.510 & 0.546 & 0.637 & 0.667 & 0.601 & 0.482 & 0.695 & 0.700 & \textbf{0.737} \\ 
bottle3 & 0.568 & 0.322 & 0.541 & 0.572 & 0.650 & 0.405 & 0.525 & 0.640 & \textbf{0.781} \\ 
bowl0 & 0.564 & 0.509 & 0.634 & 0.504 & 0.523 & 0.783 & 0.671 & 0.681 & \textbf{0.819} \\ 
bowl1 & 0.264 & 0.668 & 0.663 & 0.639 & 0.629 & 0.639 & 0.525 & 0.702 & \textbf{0.778} \\ 
bowl2 & 0.525 & 0.510 & 0.684 & 0.615 & 0.458 & 0.625 & 0.490 & 0.685 & \textbf{0.741} \\ 
bowl3 & 0.385 & 0.490 & 0.617 & 0.537 & 0.579 & 0.658 & 0.348 & 0.599 & \textbf{0.767} \\ 
bowl4 & 0.664 & 0.609 & 0.464 & 0.494 & 0.501 & 0.683 & 0.663 & 0.676 & \textbf{0.744} \\ 
bowl5 & 0.417 & 0.699 & 0.409 & 0.558 & 0.593 & 0.685 & 0.593 & \textbf{0.710} & 0.656 \\ 
bucket0 & 0.617 & 0.401 & 0.309 & 0.469 & 0.593 & 0.482 & 0.610 & 0.580 & \textbf{0.683} \\ 
bucket1 & 0.321 & 0.633 & 0.501 & 0.551 & 0.561 & 0.601 & 0.752 & \textbf{0.771} & 0.756 \\ 
cap0 & 0.668 & 0.618 & 0.557 & 0.580 & 0.589 & 0.601 & 0.693 & 0.737 & \textbf{0.822} \\ 
cap3 & 0.527 & 0.522 & 0.423 & 0.453 & 0.476 & 0.551 & 0.725 & \textbf{0.775} & 0.730 \\ 
cap4 & 0.468 & 0.520 & \textbf{0.777} & 0.757 & 0.727 & 0.553 & 0.643 & 0.652 & 0.681 \\ 
cap5 & 0.373 & 0.586 & 0.639 & \textbf{0.790} & 0.538 & 0.697 & 0.467 & 0.652 & 0.670 \\ 
cup0 & 0.403 & 0.586 & 0.539 & 0.600 & 0.610 & 0.497 & 0.510 & 0.643 & \textbf{0.776} \\ 
cup1 & 0.521 & 0.610 & 0.556 & 0.586 & 0.556 & 0.499 & 0.538 & \textbf{0.757} & \textbf{0.757} \\ 
eraser0 & 0.525 & 0.719 & 0.627 & 0.657 & 0.677 & 0.689 & 0.343 & 0.548 & \textbf{0.890} \\ 
headset0 & 0.378 & 0.520 & 0.577 & 0.583 & 0.591 & 0.643 & 0.537 & 0.720 & \textbf{0.738} \\ 
headset1 & 0.515 & 0.490 & 0.617 & 0.637 & 0.627 & 0.458 & 0.610 & 0.676 & \textbf{0.795} \\ 
helmet0 & 0.553 & 0.571 & 0.526 & 0.546 & 0.556 & 0.555 & 0.600 & 0.597 & \textbf{0.757} \\ 
helmet2 & 0.602 & 0.542 & 0.623 & 0.425 & 0.447 & 0.462 & 0.614 & \textbf{0.641} & 0.633 \\ 
helmet3 & 0.526 & 0.444 & 0.374 & 0.404 & 0.424 & 0.520 & 0.367 & 0.573 & \textbf{0.707} \\ 
helmet4 & 0.349 & 0.719 & 0.427 & 0.484 & 0.552 & 0.589 & 0.381 & 0.600 & \textbf{0.720} \\ 
jar0 & 0.420 & 0.424 & 0.441 & 0.472 & 0.483 & 0.610 & 0.592 & 0.780 & \textbf{0.838} \\ 
microphone0 & 0.563 & 0.671 & 0.357 & 0.388 & 0.488 & 0.509 & 0.414 & 0.755 & \textbf{0.762} \\ 
shelf0 & 0.164 & 0.609 & 0.564 & 0.494 & 0.523 & 0.685 & 0.688 & 0.603 & \textbf{0.696} \\ 
tap0 & 0.525 & 0.560 & 0.754 & 0.753 & 0.458 & 0.359 & 0.676 & 0.676 & \textbf{0.736} \\ 
tap1 & 0.573 & 0.546 & 0.739 & 0.766 & 0.538 & 0.697 & 0.641 & 0.696 & \textbf{0.900} \\ 
vase0 & 0.531 & 0.342 & 0.423 & 0.455 & 0.447 & 0.451 & 0.533 & 0.533 & \textbf{0.788} \\ 
vase1 & 0.549 & 0.219 & 0.427 & 0.423 & 0.552 & 0.345 & 0.702 & \textbf{0.757} & 0.729 \\ 
vase2 & 0.410 & 0.546 & 0.737 & 0.721 & 0.741 & 0.582 & 0.605 & 0.614 & \textbf{0.752} \\ 
vase3 & 0.717 & 0.699 & 0.439 & 0.449 & 0.460 & 0.582 & 0.650 & 0.700 & \textbf{0.742} \\ 
vase4 & 0.425 & 0.510 & 0.476 & 0.506 & 0.516 & 0.514 & 0.500 & 0.524 & \textbf{0.630} \\ 
vase5 & 0.585 & 0.409 & 0.317 & 0.417 & 0.579 & 0.618 & 0.520 & 0.676 & \textbf{0.757} \\ 
vase7 & 0.448 & 0.518 & 0.657 & 0.693 & 0.650 & 0.397 & 0.462 & 0.635 & \textbf{0.771} \\ 
vase8 & 0.424 & 0.668 & 0.663 & 0.662 & 0.663 & 0.529 & 0.620 & 0.630 & \textbf{0.721} \\ 
vase9 & 0.564 & 0.268 & 0.663 & 0.660 & 0.629 & 0.609 & 0.594 & 0.594 & \textbf{0.718} \\ 
\midrule
\textbf{Average} & 0.493 & 0.528 & 0.552 & 0.568 & 0.562 & 0.559 & 0.572 & 0.659 & \textbf{0.749} \\ 
\bottomrule
\end{tabular}
}
\caption{Complete image-level anomaly detection AUROC on Anomaly-ShapeNet dataset.
We highlight the best result in \textbf{bold}.}
\label{tab:shapenet-ad_all}
\end{table}

\subsubsection{Generalization on unseen data}
To assess the robustness and generalization capabilities of our proposed model, we conduct a series of experiments on different categories from diverse datasets, as outlined in Table~\ref{trans}.
The oracle result represents the performance ceiling of our model, which is obtained by training on the category that is identical to the testing.

For \textbf{known categories}, we focus on the well-regarded ShapeNetCore.v2 dataset~\cite{shapenet}, which includes categories such as Airplanes and Cars, also featured in the Real3D-AD dataset~\cite{real3dad}.
It's pertinent to note that ShapeNetCore.v2 is not an anomaly detection dataset; it does not encompass anomalous samples.
Therefore, for the first and second rows in Table~\ref{trans}, the AUROC metric cannot be utilized in this context. 
Instead, we resort to evaluating the generalization performance of models trained on Real3D-AD of the same category on ShapeNetCore.v2 using the Chamfer Distance (CD) metric.
The marked decline in performance observed upon transitioning from ShapeNetCore.v2 to Real3D-AD, and vice versa, illuminates the hurdles presented by inconsistencies between datasets.
This highlights the importance of our reconstruction approach, which effectively learns inductive biases, allowing for better generalization across different data distributions.

For \textbf{unknown categories}, we utilize the Anomaly-ShapeNet dataset~\cite{shapenetad}, as shown in the last row of Table.~\ref{trans}. 
The model was trained on a subset of Bowl and tested a category it had never encountered during training.
Remarkably, despite this lack of prior exposure, our model achieves an impressive score of 0.715 image-level AUROC.
This performance surpassed all other methods trained and tested exclusively on ``bowl4'', thus demonstrating the superior generalization capability of our method.

These results not only validate the effectiveness of our approach in handling both known and unknown categories but also underscore its potential for real-world applications where data diversity and unseen scenarios are commonplace.

\subsection{Additional main results}
Anomaly-ShapeNet~\cite{shapenetad} contains a total of 40 categories.
In Table~\ref{tab:shapenetad} of the main text, due to the space limitation, we consider objects that belong to the same kind but with differing appearances to be in the same category (e.g., bottle0, bottle1, bottle3 are categorized as Bottle).
Here, we provide the specific image-level AUROC as in Table~\ref{tab:shapenet-ad_all}.



\begin{figure}[h!]
\centering
\begin{subfigure}[b]{\linewidth}
\includegraphics[width=\linewidth]{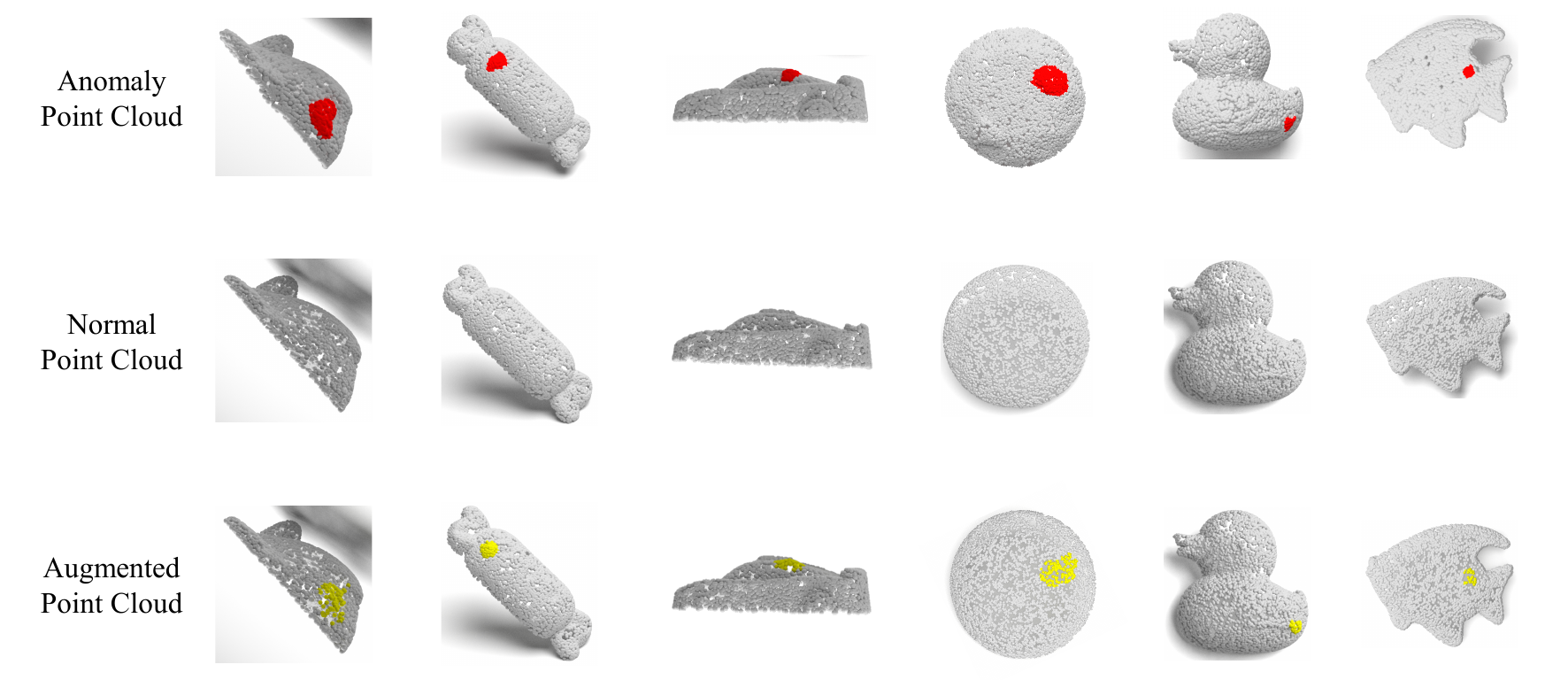}
\caption{Visualization on Real3D-AD dataset.}
\label{fig:vis_patch1}
\end{subfigure}
\begin{subfigure}[b]{\linewidth}
\includegraphics[width=\linewidth]{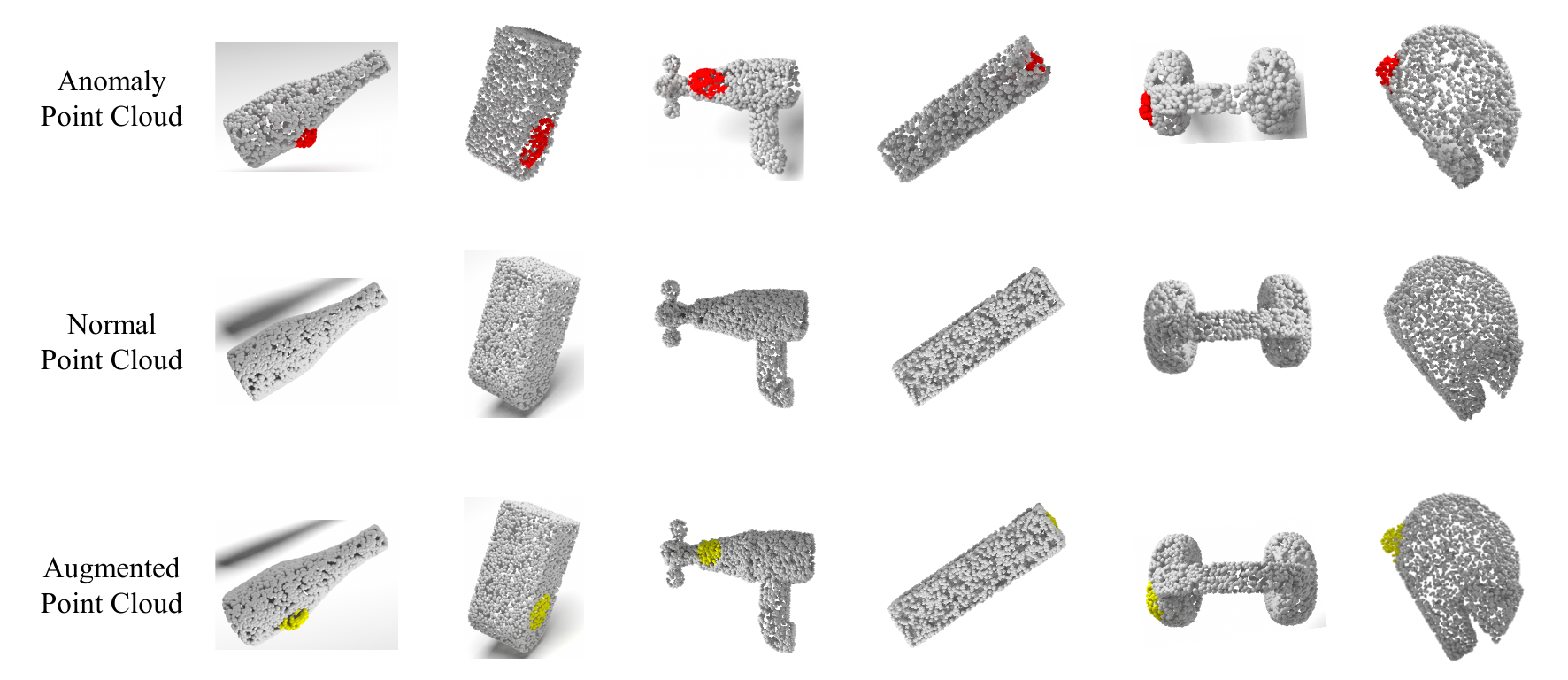}
\caption{Visualization on Anomaly-ShapeNet dataset.}
\label{fig:vis_patch2}
\end{subfigure}
\caption{Visualization on Real3D-AD dataset and Anomaly-ShapeNet dataset. The \textcolor{red}{red} region indicates the real abnormal area of the anomaly point cloud in the testing set, while the \textcolor{yellow}{yellow} region indicates the simulated abnormal area generated by Patch-Gen based on the normal point cloud in the training set.}
\label{fig:vis_patch}
\end{figure}

\subsection{Additional qualitative results}
To further demonstrate and compare the effect of our proposed 3D anomaly simulation strategy Patch-Gen, we conduct additional qualitative analysis on the Real3D-AD dataset and the Anomaly-ShapeNet dataset.

The first row shows the anomaly samples in the testing split, where the second row shows the normal samples in the training split, and the third row shows the anomaly samples simulated by Patch-Gen.
It can be seen from Fig.~\ref{fig:vis_patch} that our method fully simulates the defects that vary in different classes, proving that our method can well compensate for the domain gap caused by using only positive samples for training in 3D anomaly detection.

\end{document}